\documentclass{article}
\usepackage{spconf,amsmath,graphicx}

\usepackage{booktabs}
\usepackage{balance}

\usepackage{amsmath,epsfig,amssymb,amsthm,url,amsfonts}
\usepackage{algorithm}
\usepackage{algpseudocode}
\usepackage{multirow}
\usepackage{mdframed}
\usepackage{url}
\usepackage{enumitem}
\usepackage[makeroom]{cancel}
\usepackage{dblfloatfix}
\usepackage{array}
\usepackage{comment}
\usepackage{epstopdf}
\usepackage{float}
\usepackage{subfig}
\usepackage{breqn}
\usepackage{cite}
\usepackage{xcolor}
\usepackage{cases}
\usepackage{tabularx}
\usepackage[printonlyused,withpage]{acronym}
\usepackage{balance}
\usepackage{graphbox} 

\usepackage{hyperref}
\setlist[itemize]{label=$\triangleright$}
\newtheoremstyle{break}
{}
{}
{\itshape}
{}
{\bfseries}
{.}
{\newline}
{}
\theoremstyle{break}

\theoremstyle{definition}

\newcommand{\vect}[1]{\mathbf{#1}}
\newcommand{\bs}[1]{\boldsymbol{#1}}

\usepackage{url}

\makeatletter
\def\thmhead@plain#1#2#3{%
	\thmname{#1}\thmnumber{\@ifnotempty{#1}{ }\@upn{#2}}%
	\thmnote{ {\the\thm@notefont#3}}}
\let\thmhead\thmhead@plain
\makeatother

\graphicspath{{./figs/}}

\newsavebox\mybox

\acrodef{SE}{speech enhancement}
\acrodef{STFT}{short-time Fourier transform}
\acrodef{STOI}{short-time objective intelligibility}
\acrodef{NMF}{non-negative matrix factorization}
\acrodef{DNN}{deep neural network}
\acrodef{VAE}{variational auto-encoder}
\acrodefplural{VAEs}{variational auto-encoders}
\acrodef{MSE}{mean squared error}
\acrodef{EM}{expectation-maximization}
\acrodef{TF}{time-frequency}
\acrodef{ELBO}{evidence lower bound}
\acrodef{LR}{Living Room}
\acrodef{SDR}{signal-to-distortion ratio}
\acrodef{PESQ}{perceptual evaluation of speech quality}
\acrodef{SNR}{signal-to-noise ratio}
\acrodef{DNNs}{deep neural networks}
\acrodef{VESDE}{variance-preserving stochastic differential equation}
\acrodef{SDE}{stochastic differential equation}
\acrodef{GAN}{generative adversarial networks}
\acrodefplural{GANs}{generative adversarial networks}
\acrodef{SI-SDR}{scale-invariant signal-to-distortion ratio}
\acrodef{MOS}{mean opinion score}
\acrodef{SGMSE}{Score-based Generative Model for Speech Enhancement}
\acrodef{NCSNPP++}{Noise-Conditional Score Network}
\acrodef{WSJ}{Wall Street Journal}
\acrodef{UDiffSE}{Unsupervised Diffusion-Based Speech Enhancement}
\acrodef{PC}{Predictor-Corrector}
\acrodef{DMPS}{Diffusion Model Posterior Sampling}

\usepackage{bm}

\title{Diffusion-based Speech Enhancement with a Weighted Generative-Supervised Learning Loss}
\name{%
Jean-Eudes Ayilo, %
Mostafa Sadeghi,
Romain Serizel %
\thanks{This research was supported by the French National Research Agency (ANR) under the project REAVISE (ANR-22-CE23-0026-01). Experiments presented in this paper were carried out using the Grid'5000 testbed, supported by a scientific interest group hosted by Inria, and including CNRS, RENATER, and several universities as well as other organizations (see https://www.grid5000.fr). }}
\address{%
Université de Lorraine, CNRS, Inria, LORIA, F-54000 Nancy, France}
\begin{document}
%
\maketitle
\begin{abstract}
Diffusion-based generative models have recently gained attention in speech enhancement (SE), providing an alternative to conventional supervised methods. These models transform clean speech training samples into Gaussian noise centered at noisy speech, and subsequently learn a parameterized model to reverse this process, conditionally on noisy speech. Unlike supervised methods, generative-based SE approaches usually rely solely on an unsupervised loss, which may result in less efficient incorporation of conditioned noisy speech. To address this issue, we propose augmenting the original diffusion training objective with a mean squared error (MSE) loss, measuring the discrepancy between estimated enhanced speech and ground-truth clean speech at each reverse process iteration. Experimental results demonstrate the effectiveness of our proposed methodology.
\end{abstract}
\begin{keywords}
Speech enhancement, diffusion models, generative modeling, supervised learning.
\end{keywords}
\section{Introduction}
\label{sec:intro}


Diffusion models are a recent class of generative models that have brought significant improvements in image and audio synthesis \cite{song2021scorebased, kong2020diffwave}. Their underlying mechanism is to gradually turn training samples into noise, and then learn a parameterized model to revert this process, thus enabling data generation from pure noise. These models are also gaining increasing interest in the speech enhancement (SE) task, whose goal is to recover a clean speech signal recorded in adverse acoustic environments. In this context, diffusion models aim at learning the distribution of speech data, encoding their temporal-spectral characteristics, in order to infer clean speech from noisy observation \cite{lu2022conditional}. Additionally, this distribution is learned conditionally on the associated noisy speech data to guide the data generation process \cite{lu2022conditional,richter2022speech,hu2023noise, yen2023cold}. 

This generative-based SE approach is systematically different from the prevailing supervised counterpart \cite{xu2014regression,wang2018supervised}, which learns a \ac{DNN} to directly estimate clean speech from noisy input by minimizing a supervised loss, e.g., \ac{MSE}. In contrast, diffusion-based SE usually follows the standard unsupervised (generative-based) loss used in diffusion models, with the difference that noisy speech is provided as an additional input. While this could be potentially advantageous, as the intrinsic properties of clean speech are also modeled, contrary to supervised methods, it may not efficiently leverage the conditioned information of noisy speech. In other words, the training loss in the diffusion process does not properly measure the goodness of the estimated clean speech. As such, without a proper \textit{supervision}, it may lead to the so-called \textit{condition collapse} phenomenon \cite{tai2023revisiting}, i.e., ignoring some parts of conditioning information. 

The current study aims at addressing this issue, and bridging the performance gap between the supervised and diffusion-based approaches by combining the best of the two worlds. To this end, we propose to add an \ac{MSE} loss to the original generative-based diffusion loss. This extra supervised loss measures the distance between an estimation of the enhanced speech signal at each iteration of the reverse process and the ground-truth clean speech. In doing so, we hope to combine the effectiveness of diffusion models in unseen noise conditions and the strength of supervised methods in seen noise conditions. Experiments are performed to evaluate the effectiveness of the proposed approach against both supervised and standard diffusion-based approaches. The results show promising performance of the proposed training methodology.

In the rest of the paper, we present an overview of diffusion-based SE methods in Section~\ref{sec:sbdm}, followed by a more detailed review of \cite{richter2023speech}. Our proposed methodology is discussed in Section~\ref{eq:prop}. Next, Section~\ref{sec:exp} presents the experiments and results, and Section~\ref{sec:conc} concludes the paper.


\vspace{-2mm}
\section{Related work}\label{sec:sbdm}
\subsection{Diffusion-based speech enhancement}
The fundamental concept behind diffusion models involves two main phases. Initially, within a forward process, clean data is progressively distorted by adding (usually) Gaussian noise, eventually resulting in entirely noise data following a tractable distribution like a standard Gaussian. Subsequently, through a reverse process, a \ac{DNN} is trained to sequentially produce clean data, beginning from random noise sampled from the prior distribution.

Many recent studies have already used the diffusion principle for speech enhancement. \cite{lu2022conditional} used Gaussian Markov chains to model the forward and reverse processes, where the mean of the forward Gaussian Markov chain is a linear interpolation between the clean speech and the associated noisy one. The training objective is obtained by minimizing the Kullback–Leibler (KL) divergence between the forward and reverse Markov chains, which leads to an objective where the trained network learns to predict both Gaussian noise and non-Gaussian noise. \cite{hu2023noise} added an auxiliary classification loss to the loss function of \cite{lu2022conditional} to perform noise classification and help the model better use the noise information, which resulted in improved performance. \cite{richter2022speech} leveraged \ac{SDE} \cite{song2021scorebased} to model the forward and reverse processes, and used denoising score matching \cite{vincent2011connection} as the training objective. 

All these works condition the diffusion process on the noisy speech to take into account the non-Gaussian nature of environmental noise. More precisely, to estimate clean speech, the reverse process is performed starting from a Gaussian noise centred on the noisy speech, which allows for iteratively recovering an enhanced version. \cite{yen2023cold} proposed to extend the forward process by incorporating a deterministic, progressive degradation of the clean speech through linear interpolation between the clean speech and the noisy speech. The training objective here consists in minimizing an $\ell_1$ loss between the clean speech and the reconstruction of the clean speech at a given step of the forward degradation. The recent work \cite{tai2023revisiting} identified the condition collapse problem, and proposed an auxiliary conditional generation network for generating reliable condition representations as well as a dual-path parallel network architecture to provide fine-grained condition guidance for the diffusion model. Additionally, a refinement network is trained in a supervised way that takes the enhanced speech returned by the reverse sampling and outputs a refined version. We adopt a different strategy than these approaches by including an $\ell_2$ supervision loss in the generative denoising score matching objective.

\subsection{Score-based generative model for
SE (SGMSE)}\label{sec:sgmse}
In this section, we review the score-based diffusion model proposed in \cite{richter2022speech}, as a closely related approach to our work. Let us consider a triple of flattened \ac{STFT} representations of clean speech, noisy speech, and noise: $\mathbf{x}_0$, $\mathbf{y}$, $\mathbf{n} \in \mathbb{C}^{d}$, where $d$ denotes the total number of complex-valued \ac{TF} bins. We assume that the noisy speech \ac{STFT} is formed by the following mixture model: $\mathbf{y} = \mathbf{x}_0 + \mathbf{n}$. The objective of SE is then to recover $\mathbf{x}_0$ given $\mathbf{y}$. 

As previously mentioned, the forward process of diffusion involves gradually introducing Gaussian noise to the clean speech. This process is modeled by a stochastic differential equation (\ac{SDE}), and its solution is represented as the stochastic process $\{\mathbf{x}_t\}_t$ \cite{richter2022speech}:

\begin{equation}\label{eqn:sde_fwd}
\mathrm{d} \mathbf{x}_t=\underbrace{\gamma\left(\mathbf{y}-\mathbf{x}_t\right)}_{:=\mathbf{f}(\mathbf{x}_t, \mathbf{y})} \mathrm{d} t+\underbrace{\left[\sigma_{\textrm{min} }\left(\frac{\sigma_{\textrm{max} }}{\sigma_{\textrm{min} }}\right)^t \sqrt{\left.2 \log \left(\frac{\sigma_{\textrm{max} }}{\sigma_{\textrm{min} }}\right)\right.}\right]}_{:=g(t)} \mathrm{d} \mathbf{w}
\end{equation}
where $\mathbf{x}_t$ denotes the process state at time $t \in(0, T], \gamma \in \mathbb{R}$ controls the transition from $\mathbf{x}_0$ to $\mathbf{y}$, and $g(t) \in \mathbb{R}$, with constant parameters $\sigma_{\textrm{min}}$ and $\sigma_{\textrm{max}}$, is the diffusion coefficient that controls the amount of noise induced by a standard Wiener process $\mathbf{w}$. Moreover, $\mathbf{f}(\mathbf{x}_t, \mathbf{y})$ is the drift term, which makes the forward process conditioned on the noisy speech. For numerical stability, the forward process starts at $t_{\varepsilon}\neq 0$.

The final SE objective is to reverse the above forward process in order to estimate the clean speech. To this end, one needs to find the solution to the following associated reverse process \ac{SDE}~\cite{anderson1982}:

\begin{equation}\label{eq:rev}
\mathrm{d} \mathbf{x}_t=\left[-\mathbf{f}(\mathbf{x}_t, \mathbf{y})+g(t)^2 {\nabla_{\mathbf{x}_t} \log p_t(\mathbf{x}_t | \mathbf{y})}\right]\mathrm{d} t+g(t) \mathrm{d} \overline{\mathbf{w}}
\end{equation}
where $\overline{\mathbf{w}}$ denotes a standard Wiener process running backwards in time. In \eqref{eq:rev}, the term $\nabla_{\mathbf{x}_t} \log p_t\left(\mathbf{x}_t | \mathbf{y}\right)$ refers to the conditional score function, which is approximated by a so-called (conditional) score model $\mathbf{s}_\theta\left(\mathbf{x}_t, \mathbf{y}, t\right)$ with parameters denoted $\theta$. 
The score model can be trained by minimizing a Fisher divergence between the true and approximate score, utilising the denoising score matching principle \cite{vincent2011connection}. Doing so, the training objective reduces to:

\begin{equation}\label{eqn:train_obj_base}
{\min_{\theta} \mathbb{E}_{t,\left(\mathbf{x}_0, \mathbf{y}\right), \mathbf{z}, \mathbf{x}_t \mid\left(\mathbf{x}_0, \mathbf{y}\right)}}\Big[{L_\theta\left(\mathbf{x}_t, \mathbf{y}, t, \mathbf{z}\right)}\Big],
\end{equation}
where
\begin{equation}\label{eq:score_loss}
L_\theta\left(\mathbf{x}_t, \mathbf{y}, t, \mathbf{z}\right):=
    {\left\| \mathbf{s}_\theta\left(\mathbf{x}_t, \mathbf{y}, t\right)+\frac{\mathbf{z}}{\sigma(t)}\right\|^2}
\end{equation}
and $\mathbf{z}\sim\mathcal{N}_{\mathbb{C}}(\mathbf{z}; \bs{0}, \vect{I})$, with $\mathcal{N}_{\mathbb{C}}$ denoting the circularly-symmetric complex normal distribution. As the drift term is linear, the transition kernel $p_{0 t}\left(\mathbf{x}_t | \mathbf{x}_0, \mathbf{y}\right)$ admits a closed-form expression \cite{song2021scorebased}:

\begin{equation}\label{eq:p0t}
p_{0 t}(\mathbf{x}_t | \mathbf{x}_0, \mathbf{y})=\mathcal{N}_{\mathbb{C}}\Big(\mathbf{x}_t ; \boldsymbol{\mu}\left(\mathbf{x}_0, \mathbf{y}, t\right), \sigma(t)^2 \mathbf{I}\Big),
\end{equation}
where
\begin{equation}\label{eqn:mean}
\boldsymbol{\mu}(\mathbf{x}_0, \mathbf{y}, t)=\mathrm{e}^{-\gamma t} \mathbf{x}_0+\left(1-\mathrm{e}^{-\gamma t}\right) \mathbf{y}
\end{equation}
and
\begin{equation}\label{eqn:var-richter}
    \sigma(t)^2 = \frac{\sigma_{\textrm{min}}^{2} \Big((\sigma_{\textrm{max}}/ \sigma_{\textrm{min}})^{2t} - \textrm{e}^{-2\gamma t} \Big){\log(\sigma_{\textrm{max}}/ \sigma_{\textrm{min}})}}{\gamma + \log(\sigma_{\textrm{max}}/ \sigma_{\textrm{min}})}.
\end{equation}
Once the score model is trained, it is plugged in \eqref{eq:rev}, replacing the score function, and the resulting reverse \ac{SDE} is solved by a Predictor-Corrector sampling procedure \cite{song2021scorebased} to iteratively generate clean speech's estimates.

\section{Weighted generative-supervised learning loss}\label{eq:prop}

The training objective function described in \eqref{eqn:train_obj_base} and \eqref{eq:score_loss} aims to train the score model by minimizing the discrepancy between the approximated score and the score derived from the transition kernel, represented as $\nabla_{\mathbf{x}_t} \log p_{0 t}(\mathbf{x}_t | \mathbf{x}_0, \mathbf{y})=- {\mathbf{z}}/{\sigma(t)}$. This suggests that, in its current form, the score model is not explicitly informed of the specific SE task it is meant to perform. Instead, the training loss primarily resembles the generative (unsupervised) loss typically utilized in unconditional diffusion models.

To address this issue, we propose the inclusion of a supervised loss as a form of regularization or guidance. This additional loss explicitly reinforces the SE objective during the score model's training. To accomplish this objective, we need to have an estimate of the clean speech at each iteration of the reverse process, denoted $\hat{\mathbf{x}}_{0,t}$, to be compared against the ground-truth $\mathbf{x}_{0}$. Such an estimate could be provided using Tweedie’s approach \cite{efron2011tweedie}. To this end, by combining \eqref{eq:p0t} and \eqref{eqn:mean}, we can write 
\begin{equation}
    \mathbf{x}_t=\mathrm{e}^{-\gamma t} \mathbf{x}_0+\left(1-\mathrm{e}^{-\gamma t}\right) \mathbf{y} + \mathbf{e}_t
\end{equation}
where $\mathbf{e}_t\sim \mathcal{N}_{\mathbb{C}}(\mathbf{e}_t; \boldsymbol{0}, \sigma(t)^2\mathbf{I})$. If we apply Tweedie's formula independently to the real and imaginary parts of the variables, we obtain the following result:
\begin{align}
    \mathrm{e}^{-\gamma t}\hat{\mathbf{x}}_{0,t} + (1-\mathrm{e}^{-\gamma t})\mathbf{y}&= \mathbf{x}_t + (\sigma(t)^2/2)\nabla_{\mathbf{x}_t} \log p_t(\mathbf{x}_t | \mathbf{y})\nonumber \\
    &\approx \mathbf{x}_t + (\sigma(t)^2/2) \mathbf{s}_\theta(\mathbf{x}_t, \mathbf{y}, t).
\end{align}
We then propose to add an \ac{MSE} loss between the above expression and the associated ground-truth one, i.e., $\mathrm{e}^{-\gamma t}{\mathbf{x}}_{0} + (1-\mathrm{e}^{-\gamma t})\mathbf{y}$, to the original score loss \eqref{eqn:train_obj_base}.
This leads to the following weighted training objective for learning the conditional score model
\begin{multline}\label{eqn:train_obj_proposed}
{\min_{\theta} \mathbb{E}_{t,\left(\mathbf{x}_0, \mathbf{y}\right), \mathbf{z}, \mathbf{x}_t \mid\left(\mathbf{x}_0, \mathbf{y}\right)}}[\left(1-\alpha_t \right)  L_\theta\left(\mathbf{x}_t, \mathbf{y}, t, \mathbf{z}\right)+ \\
\alpha_t\left\| \mathbf{x}_t + \frac{\sigma(t)^2}{2}\mathbf{s}_\theta(\mathbf{x}_t, \mathbf{y}, t)   - (\mathrm{e}^{-\gamma t}{\mathbf{x}}_{0} + (1-\mathrm{e}^{-\gamma t})\mathbf{y})\right\|^2 ]
\end{multline}
which is a weighted loss between the original generative-based training loss in \eqref{eq:score_loss} and a supervised $\ell_2$ loss. $\alpha_t$ are time-dependent scalar weights taking values in $[0,1]$. We propose to use the following expression for the weights
\begin{equation}
    \alpha_t = \frac{\sigma(T)-\sigma(t)}{\sigma(T)-\sigma(t_{\varepsilon})}.
\end{equation}
In the proposed loss function \eqref{eqn:train_obj_proposed}, we aim to make a balance between two essential tasks: conditional score estimation and supervised estimation of clean speech at every reverse iteration. The parameter $\alpha_t$, as a function of time, increases as the Gaussian noise variance $\sigma(t)$ decreases during the reverse process. Consequently, at earlier stages of the reverse diffusion process, the network is expected to give higher weights to the loss associated with score estimation, while in the later stages, the score network is tasked with assigning more importance to the supervised component of the loss.

\begin{table*}[!t]
    \centering
\caption{Speech enhancement results (mean $\pm$ standard error) for WSJ0-QUT and NTCD-TIMIT under both matched and mismatched conditions. The best average metric value is highlighted in bold, and the second best is italicized.}
\resizebox{0.98\textwidth}{!}{
     \begin{tabular}{|l|l|c|c|c|c|c|c|}
    \hline
       Training set & {Metric} & {SI-SDR} (dB) & {PESQ} & {ESTOI} & {SIG-MOS} & {BAK-MOS} & {OVR-MOS} \\ \hline \hline
       \multirow{4}{*}{WSJ0-QUT} & {Input (\textbf{WSJ0-QUT})} & -2.60~$\pm$~0.16 & 1.83~$\pm$~0.02 & 0.50~$\pm$~0.01 & 4.04~$\pm$~0.01 & 2.93~$\pm$~0.02 & 3.13~$\pm$~0.01  \\  \cline{2-8} 
         & {Supervised} & \textbf{12.91~$\pm$~0.14} & 2.67~$\pm$~0.02 & \textbf{0.84~$\pm$~0.00} & 4.38~$\pm$~0.01 & \textbf{4.81~$\pm$~0.01} & 4.30~$\pm$~0.01 \\ 
         & {SGMSE+ \cite{richter2023speech}} & 10.21~$\pm$~0.16 & \textit{2.83~$\pm$~0.02} & 0.81~$\pm$~0.00 & \textit{4.52~$\pm$~0.01} & 4.70~$\pm$~0.01 & \textit{4.31~$\pm$~0.01}\\ 
        & \textbf{Proposed} & \textit{10.40~$\pm$~0.15} & \textbf{2.88~$\pm$~0.02} & \textit{0.83~$\pm$~0.00} & \textbf{4.56~$\pm$~0.01} & \textit{4.73~$\pm$~0.00} & \textbf{4.37~$\pm$~0.01} \\ \hline

         \multirow{3}{*}{NTCD-TIMIT} & {Supervised} & \textbf{9.27~$\pm$~0.15} & 2.36~$\pm$~0.02 & \textbf{0.75~$\pm$~0.00} & 4.22~$\pm$~0.01 & \textbf{4.68~$\pm$~0.01} & 4.15~$\pm$~0.02 \\ 
         & {SGMSE+ \cite{richter2023speech}} & 7.32~$\pm$~0.15 & \textit{2.51~$\pm$~0.02} & 0.72~$\pm$~0.01 & \textit{4.47~$\pm$~0.01} & 4.60~$\pm$~0.01 & \textit{4.24~$\pm$~0.01}\\ 
        & \textbf{Proposed} & \textit{7.55~$\pm$~0.14} & \textbf{2.61~$\pm$~0.02} & \textbf{0.75~$\pm$~0.00} & \textbf{4.55~$\pm$~0.01} & \textit{4.66~$\pm$~0.00} & \textbf{4.34~$\pm$~0.01} \\ \hline \hline

        \multirow{4}{*}{NTCD-TIMIT} & {Input (\textbf{NTCD-TIMIT})} & -7.81~$\pm$~0.22 & 1.77~$\pm$~0.02 & 0.31~$\pm$~0.00 & 3.51~$\pm$~0.01 & 2.28~$\pm$~0.02 & 2.69~$\pm$~0.01 \\ \cline{2-8} 
         & {Supervised} & \textbf{8.57~$\pm$~0.19} & 2.18~$\pm$~0.02 & \textit{0.54~$\pm$~0.01} & 3.69~$\pm$~0.01 & 4.26~$\pm$~0.01 & 3.42~$\pm$~0.02  \\ 
         & {SGMSE+ \cite{richter2023speech}} & 6.21~$\pm$~0.23 & \textit{2.35~$\pm$~0.02} & 0.53~$\pm$~0.01 & \textit{4.02~$\pm$~0.01} & \textit{4.30~$\pm$~0.01} & \textit{3.68~$\pm$~0.01} \\ 
        & \textbf{Proposed} & \textit{7.97~$\pm$~0.18} & \textbf{2.46~$\pm$~0.02} & \textbf{0.57~$\pm$~0.01} & \textbf{4.14~$\pm$~0.01} & \textbf{4.37~$\pm$~0.01} & \textbf{3.83~$\pm$~0.01} \\ \hline

         \multirow{3}{*}{WSJ0-QUT} & {Supervised}  & \textbf{5.98~$\pm$~0.22} & 2.02~$\pm$~0.02 & \textbf{0.50~$\pm$~0.01} & 3.76~$\pm$~0.01 & \textit{4.19~$\pm$~0.01} & 3.34~$\pm$~0.02 \\ 
         & {SGMSE+ \cite{richter2023speech}}& 1.28~$\pm$~0.27 & \textit{2.05~$\pm$~0.02} & 0.45~$\pm$~0.01 & \textit{4.04~$\pm$~0.01} & 4.05~$\pm$~0.01 & \textit{3.57~$\pm$~0.02}  \\ 
        & \textbf{Proposed} & \textit{4.42~$\pm$~0.23} & \textbf{2.08~$\pm$~0.02} & \textit{0.48~$\pm$~0.01} & \textbf{4.16~$\pm$~0.01} & \textbf{4.20~$\pm$~0.01} & \textbf{3.76~$\pm$~0.01}\\ \hline
    \end{tabular}}
     \label{tab:se_results}
\end{table*}

\section{Experiments}\label{sec:exp}


\noindent\textbf{Baselines.} We evaluate the performance of our proposed weighted generative-supervised learning loss for SE, comparing it to a reference approach referred to as SGMSE+ \cite{richter2022speech}. Both methods utilize the same network architecture, based on the Noise Conditional Score Network (NCSN++), which consists of a multi-resolution U-Net design. Additionally, we compare these two methods with a purely supervised model that directly predicts the clean speech spectrogram from the noisy speech spectrogram input. For the supervised approach, we also use the NCSN++ network, which is trained by minimizing the MSE loss between the enhanced and clean spectrograms.

\vspace{0.15cm}
\noindent\textbf{Evaluation metrics.} To measure the quality of the enhanced speech signals, we use standard instrumental evaluation metrics, including the \ac{SI-SDR} in dB \cite{le2019sdr}, the extended short-time objective intelligibility (ESTOI)  measure~\cite{jensen2016algorithm} ($[0,1]$), and the \ac{PESQ} score~\cite{rix2001perceptual} ($[-0.5,4.5]$). In addition, we use the DNS-MOS, as a non-intrusive objective evaluation metric~\cite{reddy2022dnsmos}, which provides three MOS scores: speech signal quality (SIG), background intrusiveness (BAK), and overall quality (OVRL). For all these metrics, the higher, the better. 

\vspace{0.15cm}
\noindent\textbf{Datasets.} For training and evaluation, we used the WSJ0-QUT \cite{leglaive2020recurrent} and NTCD-TIMIT datasets \cite{abdelaziz2017ntcd}, to allow for a cross-dataset evaluation. The WSJ0-QUT dataset combines the clean speech signals from the WSJ0 dataset \cite{garofolo_john_s_csr-i_2007} with noise signals from the QUT-NOISE dataset~\cite{dean2015qut}. The test subset of WSJ0-QUT comprise 651 synthetic mixtures (roughly 1.5 hours). It iscreated by taking clean speech signals from the `si\_et\_05` subset of WSJ0 (unseen speech samples) and adding noise signals sampled uniformly from the `verification` set of the QUT-NOISE dataset with \ac{SNR} values of -5, 0, and 5 dB.

The NTCD-TIMIT dataset comprises 62 English speakers (with/without Irish accent), divided into train, test, and validation subsets, where each speaker utters 98 different sentences. Duration of each utterance is approximately 5 seconds. To create the training and validation datasets, we combined the clean speech signals from the NTCD-TIMIT dataset with various types of noise from the DEMAND dataset \cite{thiemann2013diverse}. We applied different \ac{SNR} values, including -10, -5, 0, 5, and 10 dB. Each utterance in the training and validation sets was mixed with three different combinations of DEMAND noises and SNRs, resulting in a total of 12,348 training and 2,352 validation mixtures. For the test subset, we retained the same noisy speech signals as provided in the NTCD-TIMIT dataset. These noisy samples were generated by adding six different noise types, including Living Room, White, Cafe, Car, Babble, and Street, with SNRs of -5 dB, 0 dB, and 5 dB. The test set comprises 810 mixtures.


\vspace{0.15cm}
\noindent\textbf{Hyperparameters setting for SDE and STFT.} Input data representations, SDE and STFT representations follow the same settings as in \cite{richter2023speech}. Specifically, the STFT of the speech data, with a sampling rate of 16kHz, is computed with a window size of 512, a hop length of 128 (75\% overlap) and a Hann window which gives $F=256$ as the number of frequency bins. For the drift and diffusion coefficients of SDE  in ~\eqref{eqn:sde_fwd}, the parameters are set as $\gamma=1.5, \sigma_{\textrm{min}}=0.05,  \sigma_{\textrm{max}}=0.5$. The minimum and maximum process times are set to $t_{\varepsilon}=0.03$ and $T=1$, respectively. 



\vspace{0.15cm}
\noindent\textbf{Results.} In Table~\ref{tab:se_results}, we present the average speech enhancement metrics for all the cross-dataset configuration settings, along with the corresponding standard error of the mean. To clarify, we use the term ``matched condition'' when the model is trained and tested on the same dataset, while ``mismatched condition'' refers to cases where the model is evaluated on a test set from a different dataset than the one it was trained on.

From Table~\ref{tab:se_results}, we can draw several conclusions. 
First, in both matched and mismatched conditions, the supervised method consistently outperforms the two diffusion-based methods when considering the SI-SDR, ESTOI, and BAK-MOS metrics. Our proposed method ranks second in performance. Nevertheless, when trained and evaluated on the NTCD-TIMIT dataset, the ESTOI and BAK-MOS metrics show a different trend. In these cases, the supervised method underperforms our proposed method. For the PESQ, SIG-MOS, and OVR-MOS metrics, the proposed method consistently performs the best. 

A noteworthy observation from these findings is that in the matched conditions, when the supervised method performs the best, the gap between the supervised and SGMSE+ tends to narrow when the supervision loss is added. This trend also holds for the mismatched conditions.

In summary, our proposed method appears to inherit some capabilities from the supervised method, striving to match its performance in terms of SI-SDR, ESTOI, and BAK-MOS. Simultaneously, it retains and even improves upon the strengths of the baseline SGMSE+ method, resulting in better performance in terms of ESTOI, PESQ, SIG-MOS, and OVR-MOS. This suggests that the supervision loss provides valuable feedback for score estimation.

\vspace{-3.5mm}
\section{Conclusion}\label{sec:conc}
\vspace{-2mm}
In this paper, we addressed the problem of diffusion-based speech enhancement. We introduced a supervised training loss component alongside the original generative-based score estimation loss to better leverage the noisy speech data. This weighted loss balances the score estimation loss with an MSE-based supervision loss, enhancing the mapping between clean and noisy speech by incorporating clean speech estimates during the reverse process. This additional loss aids in training a score model better optimized for speech enhancement. Our experiments, conducted in both matched and mismatched conditions, demonstrate that our approach combines the strengths of supervised methods and diffusion-based approaches, resulting in improved performance. Future research directions involve exploring alternative supervised loss functions to MSE and developing more efficient adaptive weighting mechanisms.

\bibliographystyle{IEEEbib-abbrev}
\bibliography{mybib}

\end{document}